\documentclass{article}
\usepackage{spconf,amsmath,graphicx}

\usepackage{booktabs} %
\usepackage{pgfplots}
\usepackage{subcaption}
\usepackage{multirow}
\usepackage[hidelinks]{hyperref}

\usepackage{amssymb}
\usepackage{pifont}%
\newcommand{\cmark}{\ding{51}}%
\newcommand{\xmark}{\ding{55}}%

\usepackage[capitalize,noabbrev]{cleveref}

\title{Multiple Scale Latents for Learned Image Compression}
\name{Jonas Brenig, \quad Radu Timofte\thanks{This work was partially supported by the Alexander von Humboldt Foundation.}}
\address{Computer Vision Lab, CAIDAS \& IFI, University of W\"urzburg, Germany}

\usetikzlibrary{positioning,arrows,bending}
\usetikzlibrary{backgrounds}
\begin{document}

\maketitle

\begin{abstract}
Most learned image compression systems rely on a single latent representation combined with a hyperprior, which limits their ability to efficiently capture image structure across spatial scales.
In this work, we propose a hierarchical latent representation to improve the efficiency of the entropy model.
By using multiple latents at different scales, each with its own entropy model, we better capture the spatial structure of the latent representation.
Our experiments show that this approach achieves a 17.9\% BD-rate reduction over VVC on Kodak, demonstrating the effectiveness of multi-scale latent representations. %
Furthermore, the approach is orthogonal to other advances in learned image compression, making it a versatile addition to existing methods. 
\end{abstract}


\section{Introduction}
Image compression is a fundamental task in computer vision and image processing. With the increasing amount of image and video data being created, transmitted, and stored, there is a growing need for efficient and effective compression techniques.

Classical codecs such as JPEG~\cite{wallace1991jpeg}, JPEG2000~\cite{christopoulos2000jpeg2000}, and VVC~\cite{dominguez2022versatile} have long been the standard for image compression for many years. 
Although these methods are widely deployed, they rely on hand-crafted features and heuristics, whereas learned image compression methods can learn to adapt to the data distribution directly.

Recent learned image compression methods have shown promising results, outperforming traditional codecs in terms of rate-distortion performance~\cite{balle2016end, minnen2018joint, he2021checkerboard, jiang2023mlic}.
Most learned image compression methods are based on a non-linear transform coding paradigm and use an autoencoder architecture with a latent space that is compressed via an entropy model conditioned on a hyperprior~\cite{balle2015density,balle2016end,balle2018variational}.

Recent works leverage advances in deep learning architectures, such as transformers~\cite{zou2022devil,li2023frequency}, hybrid CNN-transformer models~\cite{liu2023learned,qian2022entroformer} or Mixture-of-Experts models~\cite{brenig2026moe}, to improve the performance of learned image compression methods.
Other approaches reduce redundancies in the frequency domain~\cite{fu2024weconvene} or use a correlation loss to encourage decorrelation in the latent space~\cite{ali2023towards}.

In order to further improve compression efficiency, most methods focus on improving the entropy model used for compressing the latent representation. 
Autoregressive context modeling~\cite{minnen2018joint,minnen2020channel,he2022elic} serves to minimize redundancies in the latent space.
A significant challenge in learned image compression is effectively modeling the spatial structure of the latent representation. Several works have proposed improvements to the entropy model, that aim to incorporate larger scale information~\cite{jiang2023mlic,fu2024weconvene,ali2023towards,lu2025learned,li2025learned}.

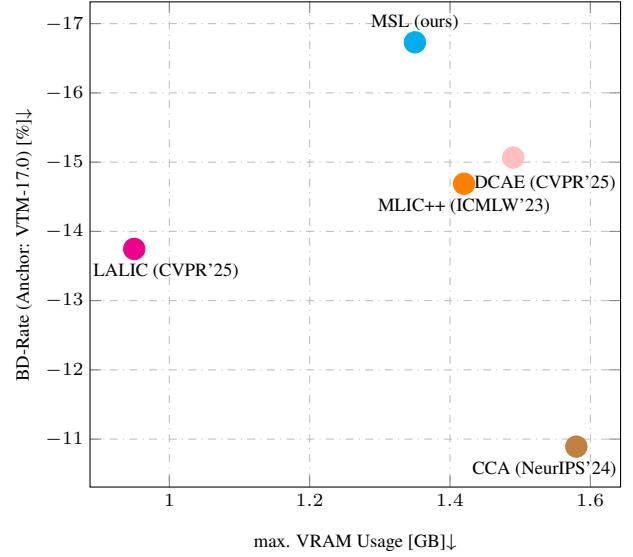
\begin{figure}[t!]
  \centering
  \begin{tikzpicture}
    \scriptsize
    \begin{axis}[
        yticklabel style={
            /pgf/number format/precision=2
        },
        xticklabel style={
            /pgf/number format/fixed,
            /pgf/number format/precision=3
        },
        scaled x ticks=false,
        xlabel=max. VRAM Usage \lbrack GB\rbrack \textdownarrow,
        ylabel style = {align=center},
        ylabel=BD-Rate (Anchor: VTM-17.0) \lbrack \%\rbrack \textdownarrow,
        ylabel near ticks,
        width=\columnwidth,
        height=8cm,
        mark size=4pt,
        grid=both,
        grid style={dashdotted},
        scatter/classes={
            mhp={cyan}, 
            dcae={pink},
            mlic={orange},
            lalic={magenta},
            elic={purple},
            tcm={blue},
            c2f={green},
            cca={brown}
        },
        y dir=reverse,
    ]        
    \addplot[
        scatter, 
        only marks,
        scatter src=explicit symbolic,
        nodes near coords*={\annotvalue },
        node near coord style={anchor=north, font=\scriptsize, yshift=\shiftvalue, xshift=\xshiftvalue},
        visualization depends on={value \thisrow{annotation} \as \annotvalue},
        visualization depends on={ value \thisrow{shift} \as \shiftvalue},
        visualization depends on={ value \thisrow{xshift} \as \xshiftvalue},
        ]  
        table[meta=label] {
            x       y                       label  annotation shift xshift
            1.35   -16.730493385443722     mhp    {MSL (ours)} +13pt 0
            1.49    -15.065656860849174     dcae   {DCAE (CVPR'25)} -4pt +12pt
            1.42   -14.687692122427476      mlic   {MLIC++ (ICMLW'23)} -3pt 0
            0.95    -13.746354022421436      lalic  {LALIC (CVPR'25)} -3pt +12pt
            1.58     -10.891394873146233     cca    {CCA (NeurIPS'24)} -3pt -12pt
        };
        
    \end{axis}
\end{tikzpicture}
  \caption{BD-Rate vs. maximum VRAM usage during decompression on the Kodak dataset~\cite{kodak1993kodak}
  }\label{fig:bd_vram}
\end{figure}

However, using a single hyper-prior may not be optimal for capturing details at different scales. 
In this work, we propose to use multiple latents at different scales to better capture the spatial structure of the latent representation.
This hierarchical latent representation allows the model to capture image features at different levels of detail, leading to improved rate-distortion performance.
Unlike prior coarse-to-fine approaches~\cite{hu2020coarse}, our model is trained end-to-end with a standard RD objective and directly integrates state-of-the-art entropy models. %
We validate our approach on standard benchmark datasets.

\section{Related Work}

\subsection{Learned Image Compression}

Most Learned Image Compression models are based on the non-linear transform coding paradigm and use an autoencoder architecture with a latent space that is compressed using an entropy model conditioned on a hyperprior~\cite{balle2016end,balle2015density,balle2018variational}. 

This approach has been extended by Minnen et al.~\cite{minnen2018joint} to use autoregressive coding, using the local context for better compression.

However, the slow fully autoregressive decoding process has been a bottleneck for practical applications.
Alternatives, such as channel-wise autoregressive coding~\cite{minnen2020channel} or checkerboard-context models~\cite{he2021checkerboard} have been proposed to address this issue. 
Recent models use channel groups~\cite{he2022elic} to further reduce the computational complexity of decoding. 

To reduce local correlations in the hyper-prior, Ali et al.~\cite{ali2023towards} proposed a correlation loss to encourage decorrelation. 
Fu et al.~\cite{fu2024weconvene} propose the use of wavelet-domain convolutional networks to reduce redundancies in the frequency domain.

In order to improve performance of autoregressive context models, Han et al.~\cite{han2024causal} propose the use of an auxiliary entropy model to provide an additional learning signal via their CCA-loss. 
Jiang et al.~\cite{jiang2023mlic,jiang2023mlic++,jiang2025mlicv2} propose an enhanced entropy model to capture local, global, and channel correlations. 
Lu et al.~\cite{lu2025learned} propose a dictionary-based entropy model to better capture different properties of the training dataset. It is based on the established channel-grouped entropy models of earlier models~\cite{he2022elic,jiang2023mlic++}, and it is further enhanced using dictionary learning for the autoregressive channel-group context prediction.

\subsubsection{Architecture}

While CNN-based methods still perform well in the field of image compression~\cite{han2024causal}, transformer-based~\cite{vaswani2017attention} methods can improve compression efficiency. %
Zou et al.~\cite{zou2022devil} and Zhu et al.~\cite{zhu2022transformer} introduced the Swin Transformer~\cite{liu2021swin} to the field of image compression.
Other works propose hybrid methods~\cite{liu2023learned,qian2022entroformer} that combine CNNs with transformers to leverage both strengths or Mixture-of-Experts models~\cite{brenig2026moe} to reduce computational requirements.
To better handle non-local information, Li et al.~\cite{li2023frequency} use frequency-aware transformer layers.

Feng et al.~\cite{feng2025linear} introduce a model using RWKV~\cite{peng2023rwkv}  as an alternative to traditional transformer blocks. Similarly, Qin et al.~\cite{qin2024mambavc} propose a Mamba~\cite{gu2024mamba}-based compression method.

\subsubsection{Multiple scale approaches}

Li et al.~\cite{li2025learned} propose a hierarchical progressive context model for learned image compression. In contrast to our work, they focus on improving the entropy model using a hierarchical structure, while we focus on the latent representation itself.

Using multiple hyper-priors at different scales has been proposed previously. Hu et al.~\cite{hu2020coarse} proposed a coarse-to-fine (C2F) hyper-prior modeling approach that uses multiple latents at different scales to capture the spatial structure of the latent representation. 
This method has shown an improved rate-distortion performance compared to single-latent methods. However, its training procedure is complex and does not incorporate recent advances in the field, such as improved context modeling.
Our approach uses a more standard training procedure and can be easily adapted to use recent advances in entropy modeling. A direct comparison is in Tab.~\ref{tab:compute}.

\section{Methodology}

\usetikzlibrary{positioning,arrows,bending}
\begin{figure*}[th]
  \centering
  \begin{subfigure}[t]{0.35\textwidth}
    \centering
    \tikzset{node distance = 4ex and 8ex}
\begin{tikzpicture}[
    module/.style={draw, very thick, rounded corners, minimum width=15ex, minimum height=4ex},
    embmodule/.style={module, fill=red!20},
    entropy/.style={module, fill=blue!20,minimum width=10ex,text width=10ex,align=center},
    mhamodule/.style={module, fill=orange!20},
    lnmodule/.style={module, fill=yellow!20},
    ffnmodule/.style={module, fill=cyan!20},
    encoder/.style={module, fill=green!20, minimum width=15ex},
    decoder/.style={module, fill=green!20, minimum width=15ex},
    projector/.style={module, fill=green!20, minimum width=7ex},
    arrow/.style={-stealth', thick, rounded corners},
    connector/.style={thick, rounded corners},
    coder/.style={module, minimum width=5ex, fill=yellow!20},
  ]
    \scriptsize
    \node (input) {$x$};

    \node (e1) [encoder,rotate=90, right= 4ex of input, anchor=north,] {Encoder $\downarrow^4$};
    \node (y1) [right= of e1.south, anchor=west] {$y$};
    \node (e2) [encoder,rotate=90, right= of y1, anchor=north] {H-Encoder $\downarrow^2$};

    \node (z)     [right=of e2.south, anchor=center] {$z$};
    \node (quant) [coder, below=11.5ex of z.center, anchor=center] {Q};
    \node (ae)    [coder, below=7ex of quant.center, anchor=center] {AE};
    \node (ad)    [coder, below=7ex of ae.center, anchor=center] {AD};
    \node (zhat)  [below=10ex of ad.center, anchor=center] {$\hat{z}$};

    \node (quant1) [coder, below=11.5ex of y1.center, anchor=center] {Q};
    \node (ae1)    [coder, below=7ex of quant1.center, anchor=center] {AE};
    \node (ad1)    [coder, below=7ex of ae1.center, anchor=center] {AD};
    \node (yhat1)  [below=10ex of ad1.center, anchor=center] {$\hat{y_0}$};

    \node (d1) [decoder,rotate=90, below=36ex of e1.center, anchor=center] {Decoder $\uparrow^4$};
    \node (d2) [decoder,rotate=90, below=36ex of e2.center, anchor=center] {H-Decoder $\uparrow^2$};

    \node (entropy) [entropy, rotate=90, above=4ex of d2.east, anchor=west] {Entropy Model};

    \node (output) [below=36ex of input.center, anchor=center] {$\hat{x}$};

    \draw[arrow] (e1) -- (y1);
    \draw[arrow] (y1) -- (e2);
    \draw[arrow] (e2) -- (z);
    \draw[arrow] (z) -- (quant);
    \draw[arrow] (quant) -- (ae);
    \draw[arrow] (ae) -- (ad);
    \draw[arrow] (ad) -- (zhat);
    \draw[arrow] (zhat) -- (d2);

    \draw[arrow] (y1) -- (quant1);
    \draw[arrow] (quant1) -- (ae1);
    \draw[arrow] (ae1) -- (ad1);
    \draw[arrow] (ad1) -- (yhat1);
    \draw[arrow] (yhat1) -- (d1);

    \draw[arrow](d2.east) -| (entropy.west);
    \draw[arrow,dotted](entropy.north) -- ++(-1.5ex,0) |- (ad1.east);
    \draw[arrow,dotted](entropy.north) -- (ae1.east);

    \draw[arrow] (input) -- (e1);
    \draw[arrow] (d1) -- (output);

    \node (size_x) [below=1ex of output.center, anchor=north] {\tiny$\left[H{\times} W\right]$};
    \node (size_y) [below=1ex of yhat1.center, anchor=north] {\tiny$\left[\frac{H}{16}{\times} \frac{W}{16}\right]$};
    \node (size_z) [below=1ex of zhat.center, anchor=north] {\tiny$\left[\frac{H}{64}{\times} \frac{W}{64}\right]$};

  \end{tikzpicture}
    \caption{Typical learned image compression method.}\label{fig:typical_architecture}
  \end{subfigure}
  \begin{subfigure}[t]{0.64\textwidth}
    \centering
    \tikzset{node distance = 4ex and 8ex}
\begin{tikzpicture}[
    module/.style={draw, very thick, rounded corners, minimum width=15ex, minimum height=4ex},
    embmodule/.style={module, fill=red!20},
    mhamodule/.style={module, fill=orange!20},
    lnmodule/.style={module, fill=yellow!20},
    ffnmodule/.style={module, fill=cyan!20},
    encoder/.style={module, fill=green!20, minimum width=15ex},
    decoder/.style={module, fill=green!20, minimum width=15ex},
    projector/.style={module, fill=green!20, minimum width=11ex},
    arrow/.style={-stealth', thick, rounded corners},
    coder/.style={module, minimum width=5ex, fill=yellow!20},
    entropy/.style={module, fill=blue!20,minimum width=10ex,text width=10ex,align=center},
  ]
    \scriptsize
    \node (input) {$x$};

    \node (e0a) [encoder,rotate=90, right=4ex of input, anchor=north] {Encoder $\downarrow$};
    \node (e0b) [encoder,rotate=90, right= of e0a.west, anchor=west] {Encoder $\downarrow$};
    \node (e1) [encoder,rotate=90, right= of e0b.west, anchor=west] {Encoder $\downarrow$};
    \node (p1) [projector,rotate=90, right= of e1.west, anchor=west] {Res Block};
    \node (e2) [encoder,rotate=90, right=9ex  of p1.west, anchor=west] {Encoder $\downarrow$};
    \node (p2) [projector,rotate=90, right= of e2.west, anchor=west] {Res Block};
    \node (e3) [encoder,rotate=90, right=9ex  of p2.west, anchor=west] {Encoder $\downarrow$};
    \node (p3) [projector,rotate=90, right= of e3.west, anchor=west] {Res Block};
    \node (e4) [encoder,rotate=90, right=9ex  of p3.west, anchor=west] {Encoder $\downarrow$};

    \node (z)     [right=6ex of e4.south, anchor=center] {$z$};

    \node (quant1) [coder, below=4ex of p1.west, anchor=center] {Q};
    \node (ae1)    [coder, below=7ex of quant1.center, anchor=center] {AE};
    \node (ad1)    [coder, below=7ex of ae1.center, anchor=center] {AD};
    \node (yhat1)  [below=10ex of ad1.center, anchor=center] {$\hat{y_2}$};

    \node (quant2) [coder, below=4ex  of p2.west, anchor=center] {Q};
    \node (ae2)    [coder, below=7ex of quant2.center, anchor=center] {AE};
    \node (ad2)    [coder, below=7ex of ae2.center, anchor=center] {AD};
    \node (yhat2)  [below=10ex of ad2.center, anchor=center] {$\hat{y_1}$};

    \node (quant3) [coder, below=4ex of p3.west, anchor=center] {Q};
    \node (ae3)    [coder, below=7ex of quant3.center, anchor=center] {AE};
    \node (ad3)    [coder, below=7ex of ae3.center, anchor=center] {AD};
    \node (yhat3)  [below=10ex of ad3.center, anchor=center] {$\hat{y_0}$};
    
    \node (quant) [coder, below=10.5ex of z, anchor=center] {Q};
    \node (ae)    [coder, below=7ex of quant.center, anchor=center] {AE};
    \node (ad)    [coder, below=7ex of ae.center, anchor=center] {AD};
    \node (zhat)  [below=10ex of ad.center, anchor=center] {$\hat{z}$};

    \node (d0a) [decoder,rotate=90, below=36ex of e0a.center, anchor=center] {Decoder $\uparrow$};
    \node (d0b) [decoder,rotate=90, below=36ex of e0b.center, anchor=center] {Decoder $\uparrow$};
    \node (d1) [decoder,rotate=90, below=36ex of e1.center, anchor=center] {Decoder $\uparrow$};
    \node (d2) [decoder,rotate=90, below=36ex of e2.center, anchor=center] {Decoder $\uparrow$};
    \node (d3) [decoder,rotate=90, below=36ex of e3.center, anchor=center] {Decoder $\uparrow$};
    \node (d4) [decoder,rotate=90, below=36ex of e4.center, anchor=center] {Decoder $\uparrow$};

    \node (entropy1) [entropy, rotate=90, above=4ex of d2.east, anchor=west] {Entropy Model};
    \node (entropy2) [entropy, rotate=90, above=4ex of d3.east, anchor=west] {Entropy Model};
    \node (entropy3) [entropy, rotate=90, above=4ex of d4.east, anchor=west] {Entropy Model};

    \node (output) [below=36ex of input.center, anchor=center] {$\hat{x}$};

    \draw[arrow] (e4) -- (z);
    \draw[arrow] (z) -- (quant);
    \draw[arrow] (quant) -- (ae);
    \draw[arrow] (ae) -- (ad);
    \draw[arrow] (ad) -- (zhat);
    \draw[arrow] (zhat) -- (d4);

    \draw[arrow] (p1) -- (quant1);
    \draw[arrow] (quant1) -- (ae1);
    \draw[arrow] (ae1) -- (ad1);
    \draw[arrow] (ad1) -- (yhat1);
    \draw[arrow] (yhat1) -- (d1);

    \draw[arrow] (p2) -- (quant2);
    \draw[arrow] (quant2) -- (ae2);
    \draw[arrow] (ae2) -- (ad2);
    \draw[arrow] (ad2) -- (yhat2);
    \draw[arrow] (yhat2) -- (d2);

    \draw[arrow] (p3) -- (quant3);
    \draw[arrow] (quant3) -- (ae3);
    \draw[arrow] (ae3) -- (ad3);
    \draw[arrow] (ad3) -- (yhat3);
    \draw[arrow] (yhat3) -- (d3);

    \draw[arrow] (d4) -- (entropy3);
    \draw[arrow] (d3) -- (entropy2);
    \draw[arrow] (d2) -- (entropy1);

    \draw[arrow, dotted] (entropy1.north) -- ++(-1ex,0) |- (ad1.east);
    \draw[arrow, dotted] (entropy1.north) -- (ae1.east);
    \draw[arrow, dotted] (entropy2.north) -- ++(-1ex,0) |- (ad2.east);
    \draw[arrow, dotted] (entropy2.north) -- (ae2.east);
    \draw[arrow, dotted] (entropy3.north) -- ++(-1ex,0) |- (ad3.east);
    \draw[arrow, dotted] (entropy3.north) -- (ae3.east);

    \draw[arrow] (input) -- (e0a);
    \draw[arrow] (d0a) -- (output);

    \draw[arrow] ([yshift=-2ex]e1.south) -- (p1.north);
    \draw[arrow] ([yshift=+5ex]e1.south) -- ([yshift=+5ex]e2.north);
    \draw[arrow] ([yshift=-2ex]e2.south) -- (p2.north);
    \draw[arrow] ([yshift=+5ex]e2.south) -- ([yshift=+5ex]e3.north);
    \draw[arrow] ([yshift=-2ex]e3.south) -- (p3.north);
    \draw[arrow] ([yshift=+5ex]e3.south) -- ([yshift=+5ex]e4.north);

    \draw[arrow] (e0a) -- (e0b);
    \draw[arrow] (d0b) -- (d0a);
    \draw[arrow] (e0b) -- (e1);
    \draw[arrow] (d1) -- (d0b);

    \node (size_x) [below=1ex of output.center, anchor=north] {\tiny$\left[H{\times} W\right]$};
    \node (size_y1) [below=1ex of yhat1.center, anchor=north] {\tiny$\left[\frac{H}{8}{\times} \frac{W}{8}\right]$};
    \node (size_y2) [below=1ex of yhat2.center, anchor=north] {\tiny$\left[\frac{H}{16}{\times} \frac{W}{16}\right]$};
    \node (size_y3) [below=1ex of yhat3.center, anchor=north] {\tiny$\left[\frac{H}{32}{\times} \frac{W}{32}\right]$};
    \node (size_z) [below=1ex of zhat.center, anchor=north] {\tiny$\left[\frac{H}{64}{\times} \frac{W}{64}\right]$};

  \end{tikzpicture}
    \caption{Using multi-scale latent variables in learned image compression.}\label{fig:architecture}
  \end{subfigure}
  \caption{Comparison of typical learned image compression architectures and our proposed multi-scale latent architecture.}\label{fig:architectures}
\end{figure*}
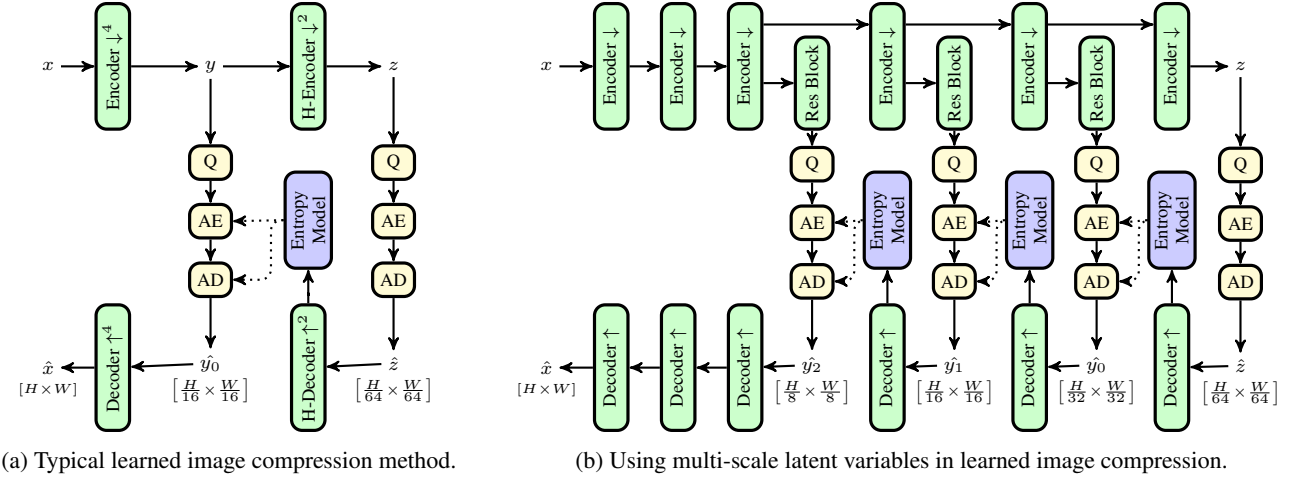

\subsection{Formulation}
Most learned image compression methods use a single hyper-prior to model the entire image, similar to the architecture shown in ~\cref{fig:typical_architecture}. Although this approach has been successful, it may not fully capture the spatial structure of the latent representation. 
Instead of using a single hyper-prior, we propose to use multiple latents at different scales to better capture image features at different levels of detail.

To achieve this, we design an architecture that uses a series of encoders and decoders to progressively compress and reconstruct the image, as shown in ~\cref{fig:architecture}. 
Each encoder and decoder operates on a different scale of the image, with the encoders progressively reducing the spatial resolution and the decoders reconstructing it. 
At every scale, we use the entropy model from \cite{lu2025learned} to compress the latents.

In principle, information can be transmitted at every scale. However, to limit computational complexity, 
in our experiments, we entropy-code and transmit latents starting from the third encoder stage.
This means that the first latent $\hat y_2$ is of shape $\frac{H}{8} \times \frac{W}{8} \times M$, the second latent $\hat y_1$ is of shape $\frac{H}{16} \times \frac{W}{16} \times M$ and the third latent $\hat y_0$ is of shape $\frac{H}{32} \times \frac{W}{32} \times M$, where $H$ and $W$ are the height and width of the input image, respectively.

In contrast, most existing methods transmit only a single latent $y$ at $\frac{H}{16} \times \frac{W}{16} \times M$ spatial resolution in addition to the hyper-latent $z$.

\subsection{Model architecture}
We implement encoders and decoders using convolutional neural networks with residual connections, as in prior work~\cite{jiang2023mlic,he2022elic}.
In order to decouple the latents $\hat y_i$ for quantization from the coarser-scale latents $y_{i-1}$, we use an additional residual block between the encoders and the entropy coders at each scale.

To compress the latents on each scale, we use the entropy model proposed by \cite{lu2025learned}, which uses dictionary-learning to better capture the properties of the training dataset.
The entropy model is based on a channel-group autoregressive model, with a latent residual prediction network.
For every slice $y_{i, j}$ of the latent $y_i$ that needs to be encoded, the model uses previously encoded slices $y_{i, <j}$ of the current scale as context, as well as $\mu$ and $\sigma$ obtained from the decoder block based on the previous scale $\hat y_{i-1}$ (or $\hat z$ for $i = 0$) to predict the distribution of the current slice. 
However, in theory, any entropy model can be used in our framework.

\subsection{Optimization}
In contrast to \cite{hu2020coarse}, who propose a complex multi-stage training procedure, we train the entire model end-to-end from scratch using a standard rate--distortion loss:
\begin{equation}
    \mathcal{L} = R + \lambda D
\end{equation}
where $R$ is the estimated bitrate, $D$ is the distortion (measured as MSE), and $\lambda$ is a hyperparameter that controls the trade-off between rate and distortion.
Following previous work~\cite{lu2025learned}, we use the straight-through estimator to handle the non-differentiable quantization operation during training.

We use latent optimization during the encoding phase based on Stochastic Gumbel Annealing (SGA), as proposed by Yang et al.~\cite{yang2020improving}, to further improve the rate-distortion performance of our model.

Due to the hierarchical nature of our latent representation, we apply SGA to each latent $y_i$ separately.
Optimizing all latents jointly resulted in unstable training and worse performance.
To reduce the optimization time, we use a reduced number of iterations of SGA for all but the highest scale latent $y_2$. 
For all smaller scale latents ($y_{i<2}, z$), we use 400 iterations of SGA with an annealing rate of $10^{-2}$. For $y_2$ we use 2000 iterations, with an annealing rate of $20^{-3}$. We use the Adam~\cite{kingma2014adam} optimizer with a learning rate of $10^{-4}$ for all latent optimization configurations.

\section{Experiments}

\subsection{Experimental Settings}

For training, we use the same dataset as Jiang et al.~\cite{jiang2023mlic}, which consists of about 100k images at 512\texttimes 512 resolution.
We train on random 256\texttimes 256 crops, using a batch size of 16 and a learning rate of $10^{-4}$ with the Adam~\cite{kingma2014adam} optimizer.
The model is trained for 2M steps. We then decay the learning rate by a factor of 10 and fine-tune for another 500k steps, followed by a final 200k steps
at a learning rate of $10^{-6}$.
To obtain results at different bitrates, we first pretrain a model for a single $\lambda = 0.013$ value and do the subsequent finetuning steps with different $\lambda$ values ranging from $\lambda = 0.0037$ to $\lambda = 0.025$.
The training is performed on Nvidia L40 GPUs.

We evaluate our model on three standard benchmark datasets, Kodak~\cite{kodak1993kodak}, CLIC~\cite{toderici2020workshop}, and Tecnick~\cite{asuni2014testimages}.
We report the rate-distortion performance in terms of PSNR vs. BPP and compare against several recent learned image compression methods, as well as the traditional codec VVC (VTM-17.0)~\cite{dominguez2022versatile}.

\subsubsection{Implementation}
Most learned image compression methods use $M=320$ channels for latent $y$ at higher quality levels, and $N=192$ channels for hyper-latent $z$.
Since we transmit information at multiple scales, we use a lower channel dimension for each latent compared to single-latent methods to limit the number of parameters of the model. For $y_i$ and $z$ we use $M=120$ channels.

The source code is available on our project page\footnote{Project page: https://jbrenig.github.io/ICIP26-MSL}.
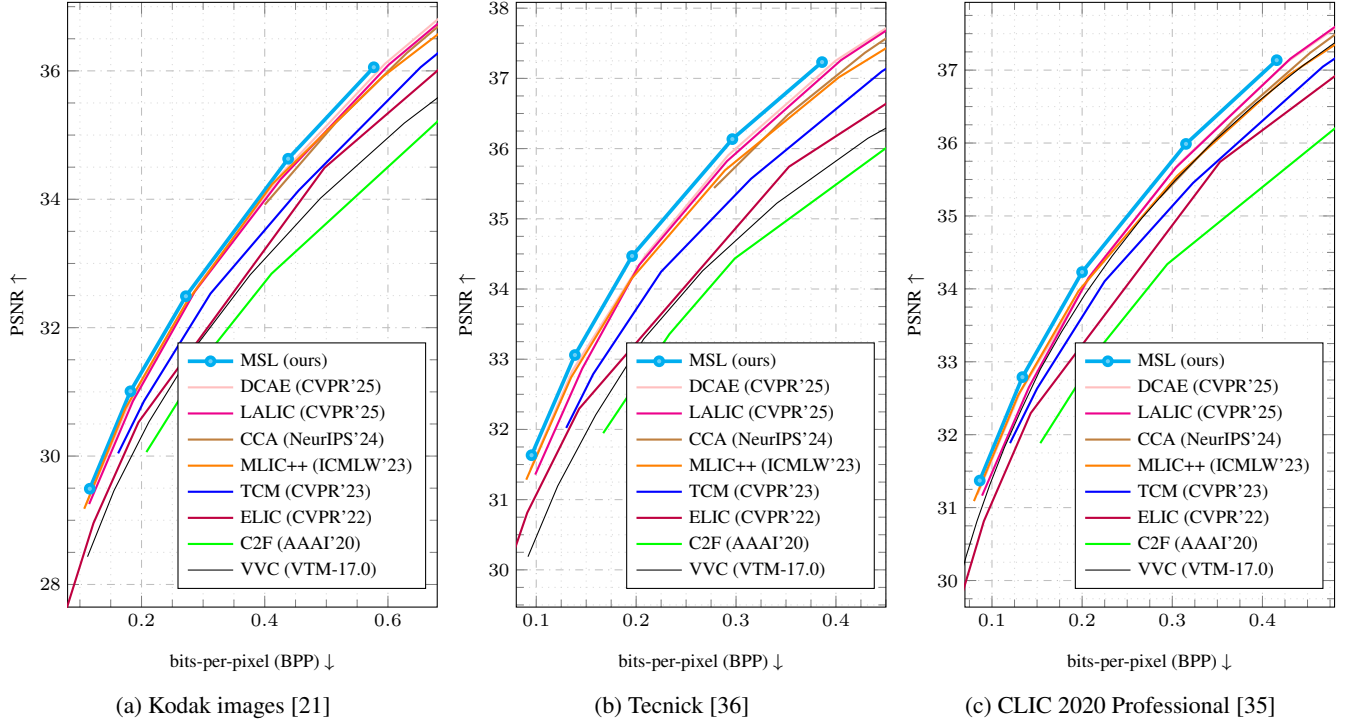
\begin{figure*}[ht]
    \pgfplotsset{
      vvc/.style={
        color=black,
        mark=none,
      },
      tcm/.style={
        color=blue,
        mark=none %
        mark options={scale=1, fill=blue!50!white},
        thick,
      },
      mlicpp/.style={ 
        color=orange,
        mark=none %
        mark options={scale=1, fill=orange!50!white},
        thick,
      },
      lalic/.style={
        color=magenta,
        mark=none %
        mark options={scale=1, fill=green!50!white},
        thick,
      },
      dcae/.style={
        color=pink,
        mark=none %
        mark options={scale=1, fill=pink!50!white},
        thick,
      },
      elic/.style={
        color=purple,
        mark=none %
        mark options={scale=1, fill=purple!50!white},
        thick,
      },
      cca/.style={
        color=brown,
        mark=none %
        mark options={scale=1, fill=brown!50!white},
        thick,
      },
      mhp/.style={
        color=cyan,
        mark=*,
        mark options={scale=1, fill=cyan!50!white},
        thick,
        line width=1.4pt
      },
      coarse2fine/.style={
        color=green,
        mark=none, %
        mark options={scale=1, fill=green!50!white},
        thick,
      },
    }
    \centering
    \scriptsize
    \begin{subfigure}[t]{0.33\textwidth}
      \centering
      \begin{tikzpicture}
    \scriptsize
    \begin{axis}[
        yticklabel style={
            /pgf/number format/precision=2
        },
        xticklabel style={
            /pgf/number format/fixed,
            /pgf/number format/precision=3
        },
        scaled x ticks=false,
        xlabel=bits-per-pixel (BPP) \textdownarrow,
        ylabel style = {align=center},
        ylabel=PSNR \textuparrow,
        ylabel near ticks,
        width=\textwidth-4em,
        height=8cm,
        mark size=1.5pt,
        grid=both,
        grid style={dashdotted},
        minor grid style={dotted,gray!30},
        minor tick num=3,
        legend cell align=left,
        legend pos=south east,
        scale only axis,
        xmin=0.08,
        xmax=0.68,
    ]

        \addplot [mhp] coordinates {  %
            (0.116, 29.491)
            (0.182, 31.009) 
            (0.272, 32.490) 
            (0.438, 34.631) 
            (0.577, 36.056) 
        }; \label{psnr_kodak_mhp}

        \addplot [dcae] coordinates {  %
            (0.110,29.243)
            (0.176,30.736)
            (0.293,32.677)
            (0.428,34.407)
            (0.595,36.109)
            (0.813,37.869)
        }; \label{psnr_kodak_dcae}

        \addplot [lalic] coordinates {  %
            (0.115,29.253)
            (0.186,30.855)
            (0.286,32.556)
            (0.426,34.312)
            (0.601,36.093)
            (0.831,37.935)
        }; \label{psnr_kodak_lalic}

        \addplot [cca] coordinates {  %
           (0.400,33.915)
           (0.515,35.189)
           (0.631,36.282)
           (0.744,37.198)
        }; \label{psnr_kodak_cca}

        \addplot [mlicpp] coordinates {  %
            (0.107,29.179)
            (0.174,30.744)
            (0.272,32.379)
            (0.412,34.235)
            (0.592,35.912)
            (0.8020,37.4594)
        }; \label{psnr_kodak_mlicpp}
        
        \addplot [tcm] coordinates {  %
            (0.162,30.042)
            (0.204,30.856)
            (0.312,32.531)
            (0.455,34.136)
            (0.653,36.056)
            (0.899,37.979)
        }; \label{psnr_kodak_tcm}
        
        \addplot [elic] coordinates {  %
            (0.044, 26.105)
            (0.074, 27.490)
            (0.122, 28.955)
            (0.195, 30.526)
            (0.497, 34.490)
            (0.865, 37.527)
        }; \label{psnr_kodak_elic}

        \addplot [coarse2fine] coordinates {  %
            (0.2079374525282118, 30.062370592702532)
            (0.30898793538411456, 31.610913728462013)
            (0.410964118109809,32.84018693415868)
            (0.7958670722113713,36.22582349988925)
            (1.0600424872504342,38.02457886151122)

        }; \label{psnr_kodak_coarse2fine}

        \addplot [vvc] coordinates {  %
            (0.1123589409722222,28.43264744879213)
            (0.155197991265191,29.45954512799564)
            (0.2119861178927952,30.53104382897368)
            (0.286176045735677,31.683147282597982)
            (0.3765860663519966,32.82541900282706)
            (0.4905403984917535,34.01671650725221)
            (0.6265335083007811,35.18614069393125)
            (0.7841169569227432,36.33389517632819)
            (0.971261766221788,37.43298582654395)
        }; \label{psnr_kodak_vvc}

        \legend{
            MSL (ours),
            DCAE (CVPR'25),
            LALIC (CVPR'25),
            CCA (NeurIPS'24),
            MLIC++ (ICMLW'23),
            TCM (CVPR'23),
            ELIC (CVPR'22),
            C2F (AAAI'20),
            VVC (VTM-17.0),
        }
    \end{axis}
\end{tikzpicture}
      \caption{Kodak images~\cite{kodak1993kodak}}\label{fig:rd_kodak}
    \end{subfigure}
    \begin{subfigure}[t]{0.33\textwidth}
      \centering
      \begin{tikzpicture}
    \scriptsize
    \begin{axis}[
        yticklabel style={
            /pgf/number format/precision=2
        },
        xticklabel style={
            /pgf/number format/fixed,
            /pgf/number format/precision=3
        },
        scaled x ticks=false,
        xlabel=bits-per-pixel (BPP) \textdownarrow,
        ylabel style = {align=center},
        ylabel=PSNR \textuparrow,
        ylabel near ticks,
        width=\textwidth-4em,
        height=8cm,
        mark size=1.5pt,
        grid=both,
        grid style={dashdotted},
        minor grid style={dotted,gray!30},
        minor tick num=3,
        legend cell align=left,
        legend pos=south east,
        scale only axis,
        xmin=0.08,
        xmax=0.45,
    ]        
        
        \addplot [mhp] coordinates {  %
            (0.095299778, 31.63298324)
            (0.1386968894, 33.05962332)
            (0.1959382218, 34.47009769)
            (0.2963053331,36.13510012)
            (0.3860851128, 37.23197044)
        }; \label{psnr_kodak_mhp}

        \addplot [dcae] coordinates {  %
            (0.094,31.430)
            (0.136,32.832)
            (0.209,34.475)
            (0.292,35.903)
            (0.400,37.255)
            (0.552,38.646)
        }; \label{psnr_kodak_dcae}

        \addplot [lalic] coordinates {  %
            (0.099,31.357)
            (0.146,32.866)
            (0.203,34.341)
            (0.291,35.822)
            (0.405,37.259)
            (0.564,38.742)
        }; \label{psnr_kodak_lalic}
        
        \addplot [cca] coordinates {  %
            (0.278, 35.439)
            (0.352, 36.482)
            (0.429, 37.365)
            (0.506, 38.107)
        }; \label{psnr_kodak_cca}

        \addplot [mlicpp] coordinates {  %
            (0.090,31.284)
            (0.135,32.743)
            (0.195,34.140)
            (0.289,35.692)
            (0.403,37.016)
            (0.553,38.331)
        }; \label{psnr_kodak_mlicpp}

        \addplot [tcm] coordinates {  %
            (0.130,32.021)
            (0.157,32.788)
            (0.225,34.246)
            (0.315,35.570)
            (0.446,37.100)
            (0.622,38.650)
        }; \label{psnr_kodak_tcm}
        \addplot [elic] coordinates {  %
            (0.034, 27.937)
            (0.056, 29.363)
            (0.091, 30.814)
            (0.143, 32.296)
            (0.353, 35.742)
            (0.627, 38.271)
        }; \label{psnr_kodak_elic}

        \addplot [coarse2fine] coordinates {  %
            (0.16700972222222213, 31.94671624977788)
            (0.23329916666666672, 33.357236653411135)
            (0.29875705555555554,34.43494854533889)
            (0.5595632777777777,37.14453189998163)
            (0.7503869999999998,38.59096453865858)

        }; \label{psnr_kodak_coarse2fine}

        \addplot [vvc] coordinates {  %
            (0.0917141111111111,30.187271359488488)
            (0.1214761666666666,31.19434031468773)
            (0.1595285555555555,32.206911844427005)
            (0.2075158333333333,33.28703546845274)
            (0.266551388888889,34.25914654249331)
            (0.3407139444444444,35.22066358622896)
            (0.4319719444444445,36.14668223341316)
            (0.5448062777777778,37.04811997240603)
            (0.6866582222222222,37.93522785041841)
        }; \label{psnr_tecnick_vvc}

        \legend{
            MSL (ours),
            DCAE (CVPR'25),
            LALIC (CVPR'25),
            CCA (NeurIPS'24),
            MLIC++ (ICMLW'23),
            TCM (CVPR'23),
            ELIC (CVPR'22),
            C2F (AAAI'20),
            VVC (VTM-17.0), 
        }
    \end{axis}
\end{tikzpicture}
      \caption{Tecnick~\cite{asuni2014testimages}}\label{fig:rd_tecnick}
    \end{subfigure}
    \begin{subfigure}[t]{0.33\textwidth}
      \centering
      \begin{tikzpicture}
    \scriptsize
    \begin{axis}[
        yticklabel style={
            /pgf/number format/precision=2
        },
        xticklabel style={
            /pgf/number format/fixed,
            /pgf/number format/precision=3
        },
        scaled x ticks=false,
        xlabel=bits-per-pixel (BPP) \textdownarrow,
        ylabel style = {align=center},
        ylabel=PSNR \textuparrow,
        ylabel near ticks,
        width=\textwidth-4em,
        height=8cm,
        mark size=1.5pt,
        grid=both,
        grid style={dashdotted},
        minor grid style={dotted,gray!30},
        minor tick num=3,
        legend cell align=left,
        legend pos=south east,
        scale only axis,
        xmin=0.07,
        xmax=0.48
    ]        

        \addplot [mhp] coordinates {  %
            (0.08640275400404523,31.370650128620426)
            (0.133746241123754,32.7894950225324)
            (0.1999414975686771,34.23018916060285)
            (0.3151841423496967, 35.98780562237995)
            (0.4158295469313133,37.13612095902606)
        }; \label{psnr_clic_mhp}
        
        \addplot [dcae] coordinates {  %
            (0.084,31.181)
            (0.131,32.575)
            (0.213,34.245)
            (0.307,35.712)
            (0.428,37.116)
            (0.596,38.580)
        }; \label{psnr_clic_dcae}

        \addplot [lalic] coordinates {  %
            (0.089,31.164)
            (0.141,32.651)
            (0.207,34.149)
            (0.304,35.664)
            (0.431,37.157)
            (0.607,38.708)
        }; \label{psnr_clic_lalic}

        \addplot [cca] coordinates {  %
            (0.286, 35.251)
            (0.368, 36.317)
            (0.453, 37.238)
            (0.538, 38.011)
        }; \label{psnr_clic_cca}
        
        \addplot [mlicpp] coordinates {  %
            (0.080,31.089)
            (0.129,32.541)
            (0.196,33.971)
            (0.305,35.545)
            (0.429,36.931)
            (0.591,38.247)
        }; \label{psnr_clic_mlicpp}

        \addplot [tcm] coordinates {  %
            (0.120,31.884)
            (0.150,32.633)
            (0.225,34.105)
            (0.323,35.452)
            (0.467,37.047)
            (0.660,38.678)
        }; \label{psnr_clic_tcm}
        
        \addplot [elic] coordinates {  %
            (0.034, 27.937)
            (0.056, 29.363)
            (0.091, 30.814)
            (0.143, 32.296)
            (0.353, 35.742)
            (0.627, 38.271)
        
        }; \label{psnr_clic_elic}

        \addplot [coarse2fine] coordinates {  %
            (0.15360779164942118, 31.883724855735792)
            (0.22436914198081379, 33.262322592937196)
            (0.29442262484609066,34.33864133026843)
            (0.5780123784763509,37.17821957878877)
            (0.7813273704568959,38.694593162078)

        }; \label{psnr_kodak_coarse2fine}

        \addplot [vvc] coordinates {
            (0.06055255, 29.8415)
            (0.07121713, 30.3267)
            (0.08331992, 30.8125)
            (0.09761736, 31.3136)
            (0.1138784, 31.8272)
            (0.13225501, 32.3358)
            (0.15360927, 32.8973)
            (0.17717163, 33.4101)
            (0.20323665, 33.9262)
            (0.23322233, 34.4499)
            (0.26638274, 34.9725)
            (0.303107, 35.4869)
            (0.34510965, 36.0125)
            (0.39253925, 36.5421)
            (0.44486286, 37.0613)
            (0.50540002, 37.5989)
            (0.57291685, 38.1335)
            (0.64778678, 38.6551)
        }; \label{psnr_clic_vvc}

        \legend{
            MSL (ours),
            DCAE (CVPR'25),
            LALIC (CVPR'25),
            CCA (NeurIPS'24),
            MLIC++ (ICMLW'23),
            TCM (CVPR'23),
            ELIC (CVPR'22),
            C2F (AAAI'20),
            VVC (VTM-17.0), 
        }
    \end{axis}
\end{tikzpicture}
      \caption{CLIC 2020 Professional~\cite{toderici2020workshop}}\label{fig:rd_clic}
    \end{subfigure}
    \caption{Rate-Distortion performance comparison on various datasets.}\label{fig:rd_performance}
\end{figure*}

\subsection{Rate-Distortion Performance}
We present the rate-distortion performance of our proposed method for various datasets in ~\cref{fig:rd_performance}.
Our method outperforms recent learned image compression methods, achieving a $-17.9\%$ BD-rate~\cite{bjontegaard2001calculation} improvement over VVC (VTM-17.0)~\cite{dominguez2022versatile} on the Kodak dataset.
For the CLIC 2020 Professional Validation dataset~\cite{toderici2020workshop} ($-9.60\%$) and the Tecnick dataset~\cite{asuni2014testimages} ($-24.1\%$) we see similar improvements in comparison to recent learned methods.

\begin{figure*}[t]
    \footnotesize
    \centering
    \input{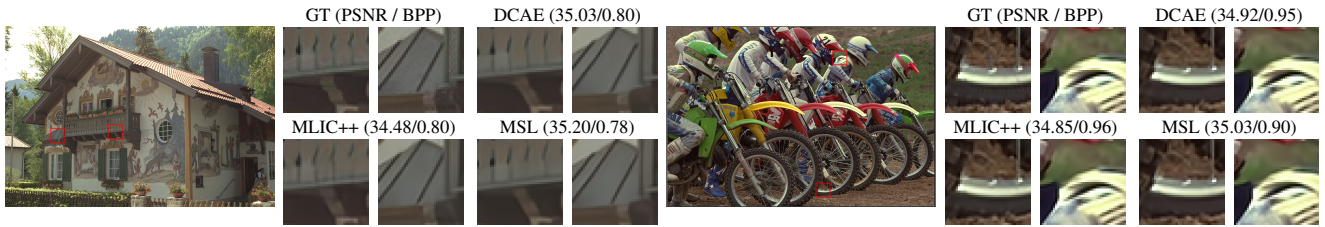}
    \caption{Visual comparison of our MSL method against recent DCAE~\cite{lu2025learned} and MLIC++~\cite{jiang2023mlic++} methods on the Kodak~\cite{kodak1993kodak} dataset. }
    \label{fig:img:compare}
\end{figure*}

In Fig.~\ref{fig:img:compare}, we present visual results comparing our method with existing learned image compression methods on the Kodak dataset.

\subsection{Computational complexity}
The multiple scale approach is inherently more computationally expensive than single scale approaches due to its autoregressive nature. Moreover, because it uses multiple entropy models, the approach uses more parameters than the baseline DCAE~\cite{lu2025learned}. 
This results in slightly slower decoding times than DCAE. 

However, due to the lower channel dimension at each scale, our approach uses less video memory, allowing the method to run on hardware with limited VRAM.

\begin{table} %
  \centering
  \begin{tabular}{lcc}
    \toprule
    Variant & BD-Rate \textdownarrow & dec (ms) \textdownarrow\\
    \midrule
    120 channels (default) & 0\%  & 115 \\
    80 channels & +4.31\% & 96 \\
    60 channels & +5.40\% & \textbf{85} \\
    \bottomrule
  \end{tabular}
  \caption{Ablation study of reduced channel dimension evaluated on the Kodak~\cite{kodak1993kodak} dataset.}\label{tab:abl:channels}
\end{table}
\begin{table*}[th]
  \centering
  \begin{tabular}{lccccc}
    \toprule
    Model & BD-Rate \textdownarrow & Params \textdownarrow & max. VRAM \textdownarrow & dec. (ms) \textdownarrow \\
    \midrule
    MSL (ours) & \textbf{-17.89\%} & 203M & 1.35GB  & 115 \\
    DCAE (CVPR'25)~\cite{li2025learned} & -15.06\% & 119M & 1.49GB& 67 \\
    MLIC++ (ICMLW'23)~\cite{jiang2023mlic++} & -14.69\% & 117M & 1.42GB & 98 \\
    LALIC (CVPR'25)~\cite{feng2025linear} & -13.74\% & 66M & 0.95GB  & 97 \\
    CCA (NeurIPS'24)~\cite{han2024causal} & -10.89\% & 65M & 0.91GB & 83 \\
    TCM (CVPR'23)~\cite{liu2023learned} & -5.43\% & 77M & 1.83GB & 93 \\
    ELIC (CVPR'22)~\cite{he2022elic} & -2.40\% & \textbf{34M} & \textbf{0.43GB} & \textbf{61} \\
    C2F~\cite{hu2020coarse} & +15.7\% & 72M & 2.30GB  & 751 \\
    \bottomrule
  \end{tabular}
  \caption{
    Quantitative comparison of learned compression methods, on the 
    Kodak~\cite{kodak1993kodak} dataset. 
    We average 5 runs, executed on a RTX 4090. VVC (VTM-17.0) serves as an anchor for BD-Rate calculation.
    }\label{tab:compute}
\end{table*}
\subsection{Ablation Study}
We perform an ablation study to analyze the impact of different design choices in our proposed method.
For this, we train several variants of our model on the same dataset for a reduced training budget of 750k steps.
\subsubsection{Number of channels M}
Since our method transmits latents at various scales, we limit the number of parameters by choosing $M=120$ channels for all scales, but other configurations are possible.
In Table~\ref{tab:abl:channels}, we evaluate the impact of this choice on the performance of the model.
Compared to a model trained with only $M=80$ channels, there is a clear improvement in rate-distortion performance with $M=120$ channels. 
We do not consider using latents with a larger channel dimension, due to significantly higher computational costs.

\subsubsection{Number of scales}
\begin{table}  %
  \centering
  \begin{tabular}{ccccccc}
    \toprule
    \multicolumn{5}{c}{Used latent scales} &\multirow{2}{*}{BD-Rate \textdownarrow} & \multirow{2}{*}{dec (ms) \textdownarrow}\\
    &$y_0$ & $y_1$ &$y_2$ & & \\
    \midrule
    &\cmark & \cmark & \cmark & & 0\% & 115\\
    &\xmark & \cmark & \cmark & & +1.95 \%  & 99\\
    &\cmark & \cmark & \xmark & & +11.60\%  & \textbf{53} \\
    \bottomrule
  \end{tabular}%
  \caption{Ablation study of using a different scale latents $y_i$ evaluated on Kodak~\cite{kodak1993kodak}. The default of using all three scales serves as the anchor for BD-Rate.}\label{tab:layers}
\end{table}

We explore the effect of varying the number of scales on the model's performance. 
We do not consider using latents with a larger spatial resolution, due to significantly higher computational costs.
We compare models using 2 scales, with either the highest resolution latent $y_2$ or the lowest resolution latent $y_0$ omitted in Tab.~\ref{tab:layers}. 
Omitting the highest resolution latent $y_2$ results in a significant drop in performance, which is likely due to the heavily reduced number of symbols available for encoding the image.
Omitting the lowest resolution latent $y_0$ results in a smaller drop in performance, but still performs worse than the full 3-scale model.

\subsubsection{Use of an additional Residual Block}

\begin{table}
  \centering
  \begin{tabular}{lcc}
    \toprule
    Variant & BD-Rate \textdownarrow & dec (ms) \textdownarrow \\
    \midrule
    with ResBlock (default)   & 0\% & 115 \\
    without ResBlock & +1.76\%  & \textbf{114} \\
    \bottomrule
  \end{tabular}
  \caption{Ablation study of additional ResBlock before quantization on the Kodak dataset.}\label{tab:abl:projection}
\end{table}

Intuitively, the additional residual Block before quantization of each latent $y_i$ serves to separate the learned representations at each scale from each other.
To verify our intuition, we compare the performance of the model with and without the inclusion of this block in Tab.~\ref{tab:abl:projection}. 
Dropping the block results in slightly worse rate-distortion performance, confirming our hypothesis.

\section{Conclusion}
In this work, we proposed a hierarchical latent representation for learned image compression using multiple latents at different scales.
Contrary to previous works with similar ideas, we train the entire model end-to-end from scratch using a standard rate-distortion loss and integrate recent advances in the field for improved context modeling.
Our experiments demonstrated that this approach leads to improved rate-distortion performance compared to single-latent methods.
The proposed method is orthogonal to other advances in learned image compression, making it a versatile addition to existing methods.
Future work could explore further architectural and training improvements, including scale-specific entropy models and reduced computational cost.

{
  \small
\bibliography{refs}

\begin{thebibliography}{10}

\bibitem{wallace1991jpeg}
Gregory~K Wallace,
\newblock ``The {JPEG} still picture compression standard,''
\newblock {\em Communications of the ACM}, vol. 34, no. 4, pp. 30--44, 1991.

\bibitem{christopoulos2000jpeg2000}
Charilaos~A Christopoulos, Touradj Ebrahimi, and Athanassios~N Skodras,
\newblock ``{JPEG2000}: the new still picture compression standard,''
\newblock in {\em 2000 ACM workshops on Multimedia}, 2000, pp. 45--49.

\bibitem{dominguez2022versatile}
Humberto~Ochoa Dominguez and Kamisetty~Ramamohan Rao,
\newblock {\em Versatile Video Coding},
\newblock River publishers, 2022.

\bibitem{balle2016end}
J.~Ballé, Valero Laparra, and Eero~P. Simoncelli,
\newblock ``End-to-end optimized image compression,''
\newblock {\em ICLR}, 2016.

\bibitem{minnen2018joint}
David Minnen, Johannes Ball{\'e}, and George~D Toderici,
\newblock ``Joint autoregressive and hierarchical priors for learned image compression,''
\newblock {\em NeurIPS}, vol. 31, 2018.

\bibitem{he2021checkerboard}
Dailan He, Yaoyan Zheng, Baocheng Sun, et~al.,
\newblock ``Checkerboard context model for efficient learned image compression,''
\newblock in {\em CVPR}, 2021, pp. 14771--14780.

\bibitem{jiang2023mlic}
Wei Jiang, Jiayu Yang, Yongqi Zhai, et~al.,
\newblock ``{MLIC}: Multi-reference entropy model for learned image compression,''
\newblock in {\em 31st ACM MM}, 2023, pp. 7618--7627.

\bibitem{balle2015density}
Johannes Ball{\'{e}}, Valero Laparra, and Eero~P. Simoncelli,
\newblock ``Density modeling of images using a generalized normalization transformation,''
\newblock in {\em 4th ICLR 2016}, 2016.

\bibitem{balle2018variational}
Johannes Ballé, David Minnen, Saurabh Singh, Sung~Jin Hwang, and Nick Johnston,
\newblock ``Variational image compression with a scale hyperprior,''
\newblock in {\em ICLR}, 2018.

\bibitem{zou2022devil}
Renjie Zou, Chunfeng Song, and Zhaoxiang Zhang,
\newblock ``The devil is in the details: Window-based attention for image compression,''
\newblock in {\em CVPR}, 2022, pp. 17492--17501.

\bibitem{li2023frequency}
Han Li, Shaohui Li, Wenrui Dai, Chenglin Li, Junni Zou, and H.~Xiong,
\newblock ``Frequency-aware transformer for learned image compression,''
\newblock {\em ICLR}, 2023.

\bibitem{liu2023learned}
Jinming Liu, Heming Sun, and Jiro Katto,
\newblock ``Learned image compression with mixed transformer-cnn architectures,''
\newblock in {\em CVPR}, 2023, pp. 14388--14397.

\bibitem{qian2022entroformer}
Yichen Qian, Xiuyu Sun, Ming Lin, Zhiyu Tan, and Rong Jin,
\newblock ``Entroformer: {A} transformer-based entropy model for learned image compression,''
\newblock in {\em The Tenth ICLR 2022}, 2022.

\bibitem{brenig2026moe}
Jonas Brenig and Radu Timofte,
\newblock ``Mixture-of-experts-based entropy model for learned image compression,''
\newblock in {\em ICIP}, 2026.

\bibitem{fu2024weconvene}
Haisheng Fu, Jie Liang, Zhenman Fang, et~al.,
\newblock ``{WeConvene}: Learned image compression with wavelet-domain convolution and entropy model,''
\newblock in {\em ECCV}. Springer, 2024, pp. 37--53.

\bibitem{ali2023towards}
Muhammad~Salman Ali, Yeongwoong Kim, Maryam Qamar, et~al.,
\newblock ``Towards efficient image compression without autoregressive models,''
\newblock {\em NeurIPS}, vol. 36, pp. 7392--7404, 2023.

\bibitem{minnen2020channel}
David Minnen and Saurabh Singh,
\newblock ``Channel-wise autoregressive entropy models for learned image compression,''
\newblock in {\em 2020 ICIP}. IEEE, 2020, pp. 3339--3343.

\bibitem{he2022elic}
Dailan He, Ziming Yang, Weikun Peng, et~al.,
\newblock ``{ELIC}: Efficient learned image compression with unevenly grouped space-channel contextual adaptive coding,''
\newblock in {\em CVPR}, 2022, pp. 5718--5727.

\bibitem{lu2025learned}
Jingbo Lu, Leheng Zhang, Xingyu Zhou, et~al.,
\newblock ``Learned image compression with dictionary-based entropy model,''
\newblock in {\em CVPR}, 2025, pp. 12850--12859.

\bibitem{li2025learned}
Yuqi Li, Haotian Zhang, Li~Li, and Dong Liu,
\newblock ``Learned image compression with hierarchical progressive context modeling,''
\newblock in {\em ICCV}, 2025, pp. 18834--18843.

\bibitem{kodak1993kodak}
Eastman Kodak,
\newblock ``Kodak lossless true color image suite (photocd pcd0992),''
\newblock {\em URL http://r0k.us/graphics/kodak}, 1993.

\bibitem{hu2020coarse}
Yueyu Hu, Wenhan Yang, and Jiaying Liu,
\newblock ``Coarse-to-fine hyper-prior modeling for learned image compression,''
\newblock in {\em AAAI}, 2020, vol.~34, pp. 11013--11020.

\bibitem{han2024causal}
Minghao Han, Shiyin Jiang, Shengxi Li, et~al.,
\newblock ``Causal context adjustment loss for learned image compression,''
\newblock {\em NeurIPS}, vol. 37, pp. 133231--133253, 2024.

\bibitem{jiang2023mlic++}
Wei Jiang, Jiayu Yang, Yongqi Zhai, et~al.,
\newblock ``{MLIC}\({}^{\mbox{++}}\): Linear complexity multi-reference entropy modeling for learned image compression,''
\newblock {\em {ACM} TOMM}, vol. 21, no. 5, pp. 142:1--142:25, 2025.

\bibitem{jiang2025mlicv2}
Wei Jiang, Yongqi Zhai, Jiayu Yang, Feng Gao, and Ronggang Wang,
\newblock ``{MLICv2}: Enhanced multi-reference entropy modeling for learned image compression,''
\newblock {\em ACM TOMM}, 2025.

\bibitem{vaswani2017attention}
Ashish Vaswani, Noam Shazeer, Niki Parmar, et~al.,
\newblock ``Attention is all you need,''
\newblock {\em NeurIPS}, vol. 30, 2017.

\bibitem{zhu2022transformer}
Yinhao Zhu, Yang Yang, and Taco Cohen,
\newblock ``Transformer-based transform coding,''
\newblock in {\em ICLR}, 2022.

\bibitem{liu2021swin}
Ze~Liu, Yutong Lin, Yue Cao, Han Hu, et~al.,
\newblock ``Swin transformer: Hierarchical vision transformer using shifted windows,''
\newblock in {\em ICCV}, 2021, pp. 10012--10022.

\bibitem{feng2025linear}
Donghui Feng, Zhengxue Cheng, Shen Wang, et~al.,
\newblock ``Linear attention modeling for learned image compression,''
\newblock in {\em CVPR}, 2025, pp. 7623--7632.

\bibitem{peng2023rwkv}
Bo~Peng, Eric Alcaide, Quentin Anthony, et~al.,
\newblock ``{RWKV}: Reinventing {RNNs} for the transformer era,''
\newblock {\em arXiv preprint arXiv: 2305.13048}, 2023.

\bibitem{qin2024mambavc}
Shiyu Qin, Jinpeng Wang, Yimin Zhou, et~al.,
\newblock ``{MambaVC}: Learned visual compression with selective state spaces,''
\newblock {\em arXiv preprint arXiv: 2405.15413}, 2024.

\bibitem{gu2024mamba}
Albert Gu and Tri Dao,
\newblock ``Mamba: Linear-time sequence modeling with selective state spaces,''
\newblock in {\em First conference on language modeling}, 2024.

\bibitem{yang2020improving}
Yibo Yang, Robert Bamler, and Stephan Mandt,
\newblock ``Improving inference for neural image compression,''
\newblock {\em NeurIPS}, vol. 33, pp. 573--584, 2020.

\bibitem{kingma2014adam}
Diederik~P. Kingma and Jimmy Ba,
\newblock ``Adam: {A} method for stochastic optimization,''
\newblock in {\em 3rd ICLR 2015}, 2015.

\bibitem{toderici2020workshop}
George Toderici, Wenzhe Shi, Radu Timofte, et~al.,
\newblock ``Workshop and challenge on learned image compression (clic2020),''
\newblock in {\em CVPR}, 2020.

\bibitem{asuni2014testimages}
Nicola Asuni, Andrea Giachetti, et~al.,
\newblock ``{TESTIMAGES}: a large-scale archive for testing visual devices and basic image processing algorithms.,''
\newblock in {\em STAG}, 2014, pp. 63--70.

\bibitem{bjontegaard2001calculation}
Gisle Bj{\o}ntegaard,
\newblock ``Calculation of average {PSNR} differences between {RD}-curves,''
\newblock 2001.

\end{thebibliography}
\bibliographystyle{IEEEbib}
}

\end{document}